\documentclass[10pt,twocolumn,letterpaper]{article}

\usepackage{iccv}
\usepackage{times}
\usepackage{epsfig}
\usepackage{graphicx}
\usepackage{amsmath}
\usepackage{caption}
\usepackage{amssymb}
\usepackage{bbm}
\usepackage{siunitx}

\newcolumntype{X}{%
  >{\rowstyle{\relax}}l%
}
\newcolumntype{Y}{%
  >{\currentrowstyle}S[detect-weight]%
}
\newcommand{\rowstyle}[1]{%
  \protected\gdef\currentrowstyle{#1}%
}

\usepackage[pagebackref=true,breaklinks=true,letterpaper=true,colorlinks,bookmarks=false]{hyperref}

 \iccvfinalcopy % *** Uncomment this line for the final submission

\def\iccvPaperID{466} % *** Enter the ICCV Paper ID here

\ificcvfinal\fi
\begin{document}

%%%%%%%%% TITLE
\title{Dense Optical Flow Prediction from a Static Image}

\author{Jacob Walker, Abhinav Gupta, and Martial Hebert\\
Robotics Institute, Carnegie Mellon University\\
{\tt\small \{jcwalker, abhinavg, hebert\}@cs.cmu.edu}
}

\maketitle
%\thispagestyle{empty}

%%%%%%%%% ABSTRACT
\begin{abstract}
Given a scene, what is going to move, and in what direction will it move?
Such a question could be considered a non-semantic form of action prediction.
In this work, we present a convolutional neural network (CNN) based approach for motion prediction. Given a 
static image, this CNN predicts the future motion of each and every pixel in the image
in terms of optical flow. Our CNN model leverages the data in tens of thousands 
of realistic videos to train our model. Our method relies on absolutely no human labeling
and is able to predict motion based on the context of the scene. Because our CNN model makes no assumptions 
about the underlying scene, it can predict future optical flow on a diverse set of scenarios. We outperform all previous approaches by large margins.
\end{abstract}

%%%%%%%%% BODY TEXT
\vspace{-0.2in}
\section{Introduction}
\vspace{-0.1in}
Consider the images shown in Figure~\ref{teaser}. Given the girl in front of the cake, we humans can easily predict that her head will move downward to extinguish the candle. The man with the discus is in a position to twist his body strongly to the right, and the squatting man on the bottom has nowhere to move but up. Humans have an amazing ability to not only recognize what is present in the image but also predict what is going to happen next. Prediction is an important component of visual understanding and cognition. In order for computers to react to their environment, simple activity detection is not always sufficient. For successful interactions, robots need to predict the future and plan accordingly. 

\setlength{\tabcolsep}{0.0pt}
\begin{figure}
\begin{tabular}{ ccc }
\includegraphics[width=0.25\textwidth]{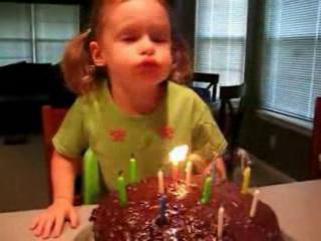} &
\includegraphics[width=0.25\textwidth]{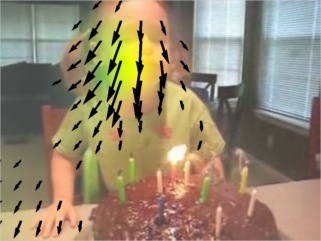} &
\\
\includegraphics[width=0.25\textwidth]{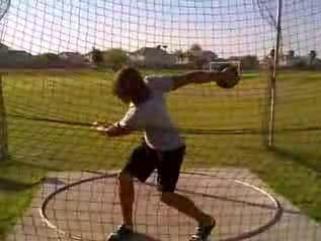} &
\includegraphics[width=0.25\textwidth]{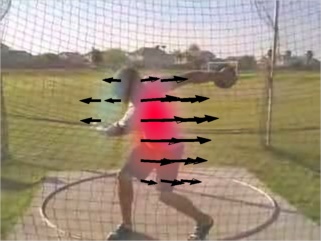} &
\\
\includegraphics[width=0.25\textwidth]{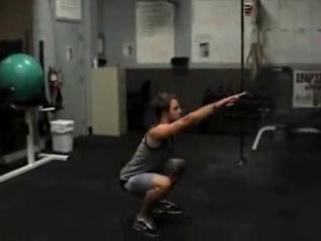} &
\includegraphics[width=0.25\textwidth]{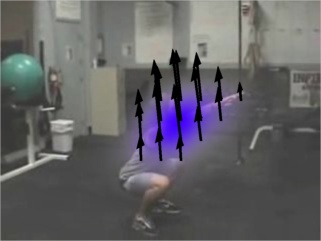} &
\\

{\footnotesize (a) Input Image} & {\footnotesize (b) Prediction} \\
\end{tabular}  
\setlength{\tabcolsep}{10pt}
\begin{tabular}{ p{5cm} c }
\vspace{-0.1in}
\caption{\small \textbf{Motion Prediction.}  Consider single, static input images (a). Our method can first identify what these actions are and predict (b) correct motion based on the pose and stage of the 
action without any video information. We use the color coding from ~\cite{MiddleBuryCode} shown
on the right.} \label{teaser} &
\raisebox{-1.2\height}{\protect\includegraphics[scale=0.8]{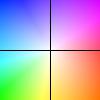}}
\end{tabular}{}
  \vspace{-1.5cm}
\end{figure}

There has been some recent work that has focused on this task. The most common approach to this prediction problem is to use a planning-based agent-centric approach: an object~\cite{Kitani12} or a patch~\cite{Walker14} is modeled as an agent that performs actions based on its current state and the goal state. Each action is decided based on compatibility with the environment and how these actions helps the agent move closer to the goal state. The priors on actions are modeled via transition matrices. Such an approach has been shown to produce impressive results: predicting trajectories of humans in parking lots~\cite{Kitani12} or hallucinating car movements on streets~\cite{Walker14}.  There are two main problems with this approach. First, the predictions are still sparse, and the motion is still modeled as a trajectory. Second, and more importantly, these approaches have always been shown to perform in restrictive domains such as parking lots or streets.

In this paper, we take the next step towards generalized prediction --- a  framework that can be learned from tens of thousands of realistic videos. This framework can work in indoor and outdoor environments; it can account for one or multiple agents whether the agent is an animal, a human, or even a car. Specifically, this framework looks at the task of motion prediction --- given a static image we predict the dense expected optical flow as if this image were part of a video. This optical flow represents how and where each and every pixel in the image is going to move in the future. However, we can see that motion prediction is more than identifying active agents; it is also highly dependent on context. For example, someone's entire body may move up or down if they are jump-roping, but most of the body will be stationary if they are playing the flute. Instead of modeling agents and its context separately under restrictive assumptions, we use a learning based approach for motion prediction. Specifically, we train a deep network that can incorporate all of this contextual information to make accurate predictions of future motion in a wide variety of scenes. We train our model from thousands of realistic video datasets, namely UCF-101 ~\cite{Soomro12} and the HMDB-51 ~\cite{Kuehne11}.

\noindent {\bf Contributions:} Our paper makes three contributions. First, we present a CNN model for motion prediction. Given a static image, our CNN model predicts expected motion in terms of optical flow. Our CNN-based model is agent-free and makes almost no assumptions about the underlying scene. Therefore, we show experimental results on diverse set of scenes. Second, our CNN model gives state of the art performance on prediction compared to contemporary approaches. Finally, we also present a proof of concept extension of the CNN model which makes long-range prediction about future motion. Our preliminary results indicate that this new CNN model might indeed be promising even for the task of long-range prediction.

\vspace{-0.1in}
\section{Background}
\vspace{-0.1in}
Prediction has caught the interest of the vision community in recent years. Most of research in this area has looked at different aspects of the problem. The first aspect of interest is the output space of prediction. Some of the initial work in this area focused on predicting the trajectory for the given input image~\cite{Yuen10}. Others have looked at more semantic forms of prediction: that is, predicting the action class of what is going to happen next~\cite{Hoai14, Savarese14}. However, one of the issues with semantic prediction is that it tells us nothing about the future action beyond the 
category. One of our goals in prediction is to go beyond classification and predict the spatial layout of future actions. For example, in case of agents such as humans, the output space of prediction can be trajectories themselves~\cite{Kitani12}. On the contrary, recent approaches have argued for much richer form of predictions even in terms of pixels~\cite{Walker14} or the features of the next frame~\cite{Kitani14, Ranzato14}.

The other aspect of research in visual prediction looks at the question of selecting the right approach for prediction. There have been two classes of approaches for the temporal prediction. The first is a data-driven, non-parametric approach. In the case of non-parameteric approaches, they do not make any assumptions about the underlying scene. For example, ~\cite{Yuen10} simply retrieves videos visually similar to the static scene, allowing a warping ~\cite{Liu11} of the matched action into the scene. The other end of the spectrum is parametric and domain-specific approaches. Here, we make assumptions on what are the active elements in the scene whether they may be cars or people. Once the assumption is made, then a model is developed to predict agent behavior. This includes forecasting pedestrian trajectories~\cite{Kitani12}, human-human interactions ~\cite{Kitani14, Savarese14}, human expressions through SOSVM~\cite{Hoai14}, and human-object interaction through graphical models ~\cite{Saxena13, Fouhey14}.

Some of the recent work in this area has looked at a more hybrid approach. For example, Walker et al. ~\cite{Walker14} builds a data-derived dictionary of rigid objects given a video domain and then makes long-term motion and appearance predictions using a transition and context model. Recent approaches such as ~\cite{Ranzato14} and ~\cite{Srivastava15} have even looked at training convolutional neural networks for predicting one future frame in a clip~\cite{Ranzato14} or motion of handwritten characters~\cite{Srivastava15}.

We make multiple advances over previous work in this paper. First, our self-supervised method can generalize across a large number of diverse domains. While ~\cite{Walker14} does not explicitly require video labels, it is still domain dependent, requiring a human-given distinction between videos in and outside the domain. In addition, ~\cite{Walker14} focused only on birds-eye domains where scene depth was limited or non existent, while our method is able to generalize to scenes with perspective.
~\cite{Pintea14} also uses  self-supervised methods to train a Structured Random Forest for motion prediction. However, the authors only learn a model from the simple KTH ~\cite{KTHdata} dataset. We show that our method is able to learn from a set of videos that is far more diverse across scenes and actions. In addition, we demonstrate much better generalization can be obtained as compared to the nearest-neighbor approach of Yuen et al.~\cite{Yuen10}. 

\noindent {\bf Convolutional Neural Networks:} We show in this paper that a convolutional neural network can be trained for the task of motion prediction in terms of optical flow. Current work on CNNs have largely focused on recognition tasks both in images and video ~\cite{Krizhevsky12, Karpathy14, Girshick14, Szegedy14, Zhang13, Simonyan14, Srivastava15}. There has been some initial work where CNNs have been combined with recurrent models for prediction. For example, ~\cite{Ranzato14} uses a LSTM ~\cite{LSTM} to predict the immediate next frame given a video input.  ~\cite{Srivastava15} uses a recurrent architecture to predict motions of handwritten characters from a video. On the other hand, our approach predicts motion for each and every pixel from a static image for any generic scene.

%Unlike these papers, our work specifically focuses on prediction as a scene understanding problem --- we use CNNs to predict future motion in a single scene. We do, however, extend our network to be "deep" in time to predict motion over a series of multiple frames. Our architecture is similar to ~\cite{Krizhevsky12} with some modifications. However, we emphasize that we can train our network in an unsupervised manner by using optical flow as a label. We find that the representation learned by our network can perform on a level comparable to pretrained, supervised features. 

%~\cite{Walker14} showed that the mid-level HOG patch framework ~\cite{Singh12},~\cite{Doersch12},~\cite{Doersch13} combined with a learned motion model can be used to identify active elements in the scene in an unsupervised way. However, this framework is domain-dependent, requiring a negative dataset to train against the given domain. ~\cite{Pintea14} had similar goals to ours, and their SRF method was unsupervised. However, we find that their implementation is very difficult to scale to large datasets beyond those such as the KTH  ~\cite{KTHdata} dataset. 

%\section{Overview}
%\vspace{-0.05in}
\begin{figure*}
\centering
\includegraphics[width=6.5in, height=2in]{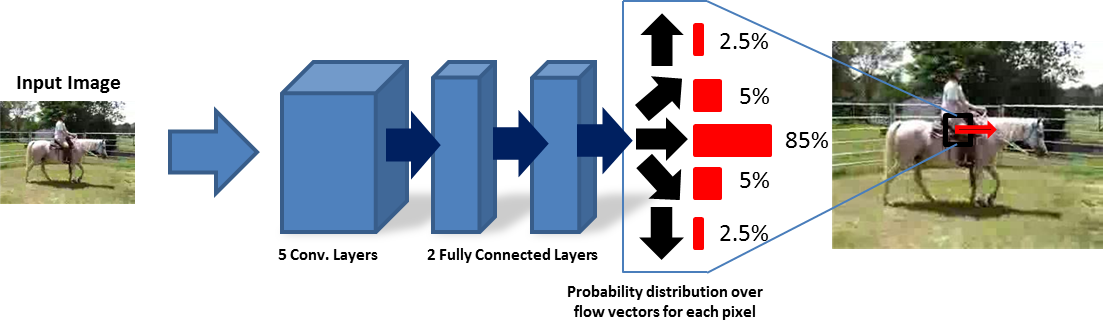} 
\vspace{-0.2in}
\caption{\textbf{Overview}. \small Our network is similar to the standard 7-layer architecture ~\cite{Krizhevsky12} used for many recognition tasks. We take a 200x200 image as input. However, we use a spatial softmax as the final output. For every pixel in the image we predict a distribution of various motions with various directions and magnitudes. We can combine a weighted average of these vectors to produce the final output for each pixel. For computational reasons, we predict a coarse 20x20 output.}
\vspace{-0.2in}
\label{architecture}
\end{figure*}

%Our goal is to predict the immediate future motion of active objects given the context of the scene. We also wish to make our method as generalizable as possible --- we desire no necessary human labeling. 

\vspace{-0.1in}
\section{Methods}
\vspace{-0.1in}
\setlength{\tabcolsep}{1pt}
\begin{figure}
\begin{tabular}{ ccc }
\includegraphics[width=0.16\textwidth]{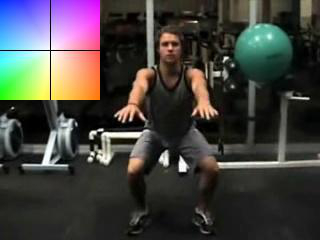} &
\includegraphics[width=0.16\textwidth]{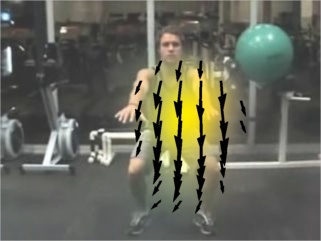} &
\includegraphics[width=0.16\textwidth]{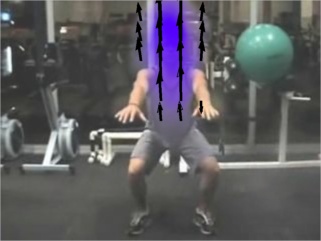} \\
\includegraphics[width=0.16\textwidth]{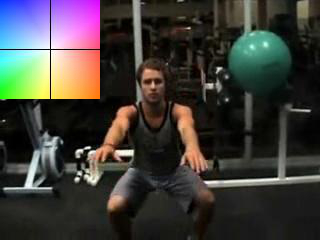} &
\includegraphics[width=0.16\textwidth]{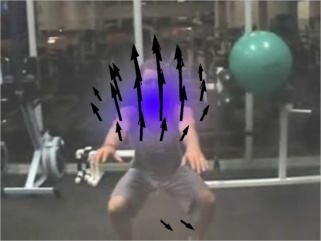} &
\includegraphics[width=0.16\textwidth]{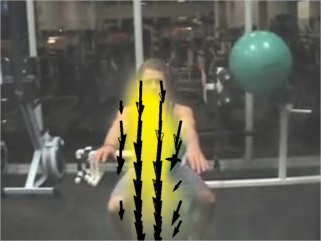} \\
{\footnotesize (a) Input Image} & {\footnotesize (b) Prediction} & {\footnotesize (c) Ground Truth}\\
\end{tabular}
\setlength{\tabcolsep}{10pt}
\vspace{-0.1in}
\caption{\small Consider the images on the left. Is the man squatting up or down? The
bottom is near completion (or just starting), and the top image is right in the
middle of the action. Our dataset contains a large number of ambiguous images such 
as these. In our evaluation we consider the underlying distribution of movements
predicted by our network. It is highly likely that this man is going to move up 
or down, but unlikely that he will veer off to the left or right.} \label{labelvariability} 
\vspace{-0.2in}
\end{figure}

Our goal is to learn a mapping between the input RGB image and the output space which corresponds to the predicted motion of each and every pixel in terms of optical flow. We propose to use CNNs as the underlying learning algorithm for this task. However, there are a few questions that need to be answered: what is a good output space, and what is a good loss function? Should we model optical flow prediction as a regression or a classification problem? What is a good architecture to solve this problem? We now discuss these issues below in detail.

\vspace{-0.1in}
\subsection{Regression as Classification} 
\vspace{-0.05in}
Intuitively, motion estimation can be posed as a regression problem since the space is continuous. Indeed, this is exactly the approach used in ~\cite{Pintea14}, where the authors used structured random forests to regress the magnitude and direction of the optical flow. However, such an approach has one drawback: such an output space tends to smoothen results to the mean. Interestingly, in a related regression problem of surface normal prediction, researchers have proposed reformulating structured regression as a classification problem ~\cite{Wang15, Ladicky14}. Specifically, they quantize the surface normal vectors into a codebook of clusters and then output space becomes predicting the cluster membership. In our work, we take a similar approach. We quantize optical flow vectors into $40$ clusters by k-means. We can then treat the problem in a manner similar to semantic segmentation, where we classify each region as the image as a particular cluster of optical flow. We use a soft-max loss layer at the output for computing gradients.

However, at test time, we create a soft output by considering the underlying distribution of all the clusters, taking a weighted-probability sum over all the classes in a given pixel for the final output. Transforming the problem into classification also leads directly to a discrete probability distribution over vector directions and magnitudes. As the problem of motion prediction can be ambiguous depending on the image (see Figure ~\ref{labelvariability}), we can utilize this probability distribution over directions to measure how informative our predictions are. We may be unsure if the man in Figure ~\ref{labelvariability} is sitting down or standing up given only the image, but we can be quite sure he will not turn right or left. In the same way,  our network can rank upward and downward facing clusters much higher than other directions. Even if the ground truth is upward, and the highest ranked cluster is downward, it may be that the second-highest cluster is also upward. Because the receptive fields are shared by the top layer neurons, the output trends to a globally coherent movement. A discrete probability distribution, through classification, allows an easier understanding of how well our network may be performing. 

\vspace{-0.1in}
\subsection{Network Design} 
\vspace{-0.05in}
Our model is similar to the standard seven-layer architecture from ~\cite{Krizhevsky12}. To simplify the description,
we denote the convolutional layers as $C(k,s)$, which indicates that there are $k$ kernels, each having the size of
$s \times s$. During convolution, we set all the strides to $1$ except for the first layer, which is $4$. We also denote the local response normalization layer as LRN,
and the max-pooling layer as MP. The stride for pooling is $2$ and we set the pooling operator size as $3 \times 3$. Finally, $F(n)$ denotes fully connected layer with $n$ neurons. Our network architecture can be described as:

 $C(96,11) \rightarrow LRN \rightarrow P \rightarrow C(256, 5) \rightarrow LRN \rightarrow P \rightarrow C(384,3) \rightarrow C(384,3) \rightarrow C(256,3) \rightarrow P \rightarrow F(4096) \rightarrow F(4096)$. We used a modified version of the popular Caffe toolbox ~\cite{caffe} for our implementation. For computational simplicity, we use 200x200 windows as input.  We used a learning rate of 0.0001 and a stepsize of 50000 iterations. Other network parameters were set to default. The only exception is that we used Xavier initialization of parameters. Instead of using the default softmax output, we used a spatial softmax loss function from ~\cite{Wang15} to classify every region in the image. This leads to a $M\times N\times C$ softmax layer, where $M$ is the number of rows, $N$ is the number of columns, and $C$ is the number of clusters in our codebook. We used $M=20$, $N=20$, and $C=40$ for a softmax layer of 16,000 neurons. Our softmax loss is spatial, summing over all the individual region losses. Let $I$ represent the image and $Y$ be the ground truth optical flow labels represented as quantized clusters. Then our spatial loss function $L(I,Y)$ is:

\begin{equation}
L(I,Y) = - \sum_{i=1}^{M\times N}\sum_{r=1}^{C}(\mathbbm{1}(y_{i} = r) \log{F_{i,r}(I))}
\end{equation}

where $F_{i,r}(I)$ represents the probability that the $i$th pixel will move according to cluster $r$. $\mathbbm{1}(y_{i} = r)$ is
an indicator function. 

\textbf{Data Augmentation:} For many deep networks, datasets which are insufficiently diverse or too small will lead to overfitting. ~\cite{Simonyan14} and ~\cite{Karpathy14} show that training directly on datasets such as the UCF-101 for action classification leads to overfitting, as there is only on the order of tens of thousands of videos in the dataset. However, our problem of single-frame prediction is different from this task. We find that we are able to build a generalizable representation for prediction by training our model over 350,000 frames from the UCF-101 dataset as well as over 150,000 frames from the HMDB-51 dataset. We benefit additionally from data augmentation techniques. We randomly flip each image as well as use randomly cropped windows. For each input, we also mirror or flip the respective labels. In this way we are able to avoid spatial biases (such as humans always appearing in the middle of the image) and train a general model on a far smaller set of videos than for recognition tasks.

\textbf{Labeling:} We automatically label our training dataset with an optical flow algorithm. We chose the publicly available implementation of DeepFlow ~\cite{Weinzaepfel13} to compute optical flow. The UCF-101 and the HMDB-51 dataset use realistic, sometimes low-quality videos from a wide variety of sources. They often suffer from compression artifacts. Thus, we aim to make our labels somewhat less noisy by taking the average optical flow of five future frames for each image. The videos in these datasets are also unstabilized. ~\cite{Wang13} showed that action recognition can be greatly improved with camera stabilization. In order to further denoise our labels, we wish to focus on the motion of objects inside the image, not the camera motion. We thus use the stabilization portion of the implementation of ~\cite{Wang13} to automatically stabilize videos using an estimated homography. 

%is thus more likely to overfit. However, this effect is mitigated by noisy labels as well as the general ambiguity in predicting motion in a single frame. In addition, we found that using spatial pixel level prediction lead to regression to the mean. Thus, we clustered our optical flow frames into a codebook of 1000 frames, and we posed the problem as a sequential prediction problem similar to caption generation. Instead of a sequence of words, our "words" are clusters of optical flow frames, and our "sentence" is an entire trajectory. We used a set number of sequences, six, in our experiments with each frame representing the average optical flow of one-sixth of a second. 

\vspace{-0.1in}
\section{Experiments}
\vspace{-0.1in}

\begin{figure*}
\centering
\begin{tabular}{ cccccccc }
\includegraphics[width=0.12\textwidth]{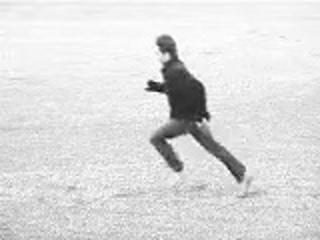} &
\includegraphics[width=0.12\textwidth]{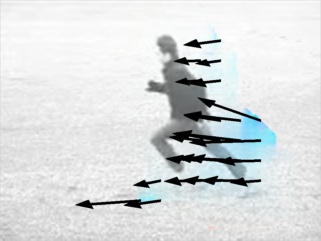} &
\includegraphics[width=0.12\textwidth]{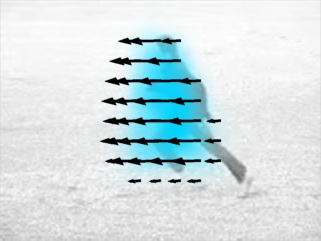} &
\includegraphics[width=0.12\textwidth]{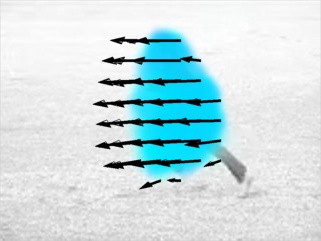} &
\includegraphics[width=0.12\textwidth]{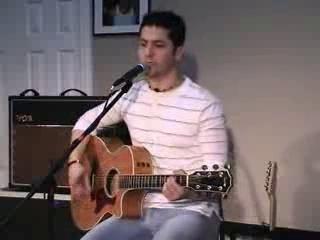} &
\includegraphics[width=0.12\textwidth]{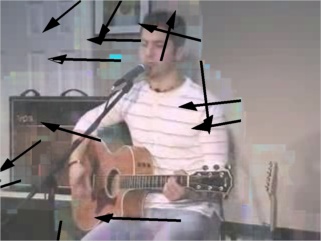} &
\includegraphics[width=0.12\textwidth]{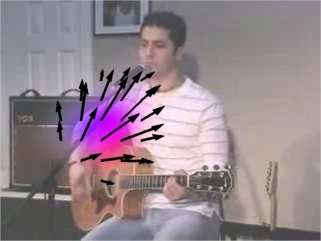} &
\includegraphics[width=0.12\textwidth]{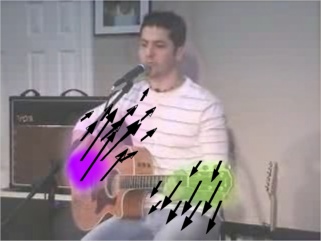} \\
\includegraphics[width=0.12\textwidth]{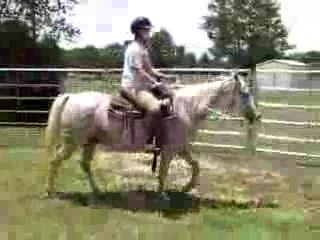} &
\includegraphics[width=0.12\textwidth]{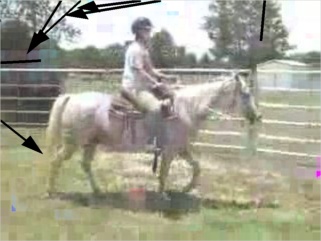} &
\includegraphics[width=0.12\textwidth]{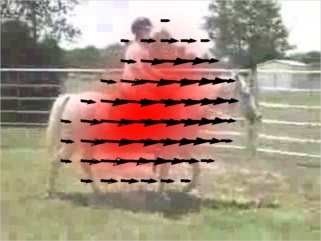} &
\includegraphics[width=0.12\textwidth]{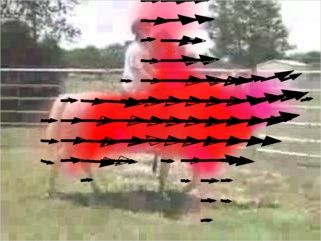} &
\includegraphics[width=0.12\textwidth]{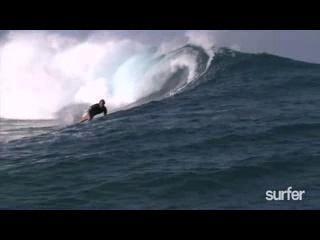} &
\includegraphics[width=0.12\textwidth]{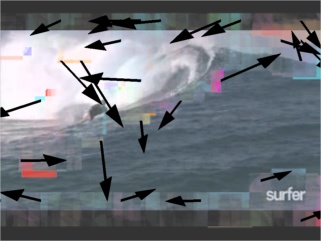} &
\includegraphics[width=0.12\textwidth]{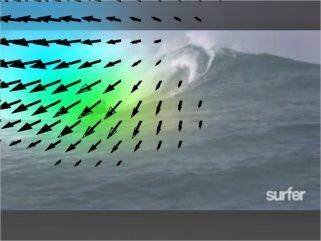} &
\includegraphics[width=0.12\textwidth]{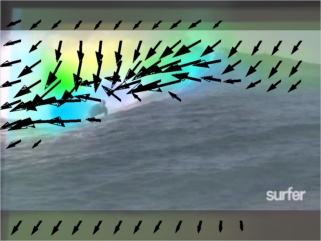} \\
\includegraphics[width=0.12\textwidth]{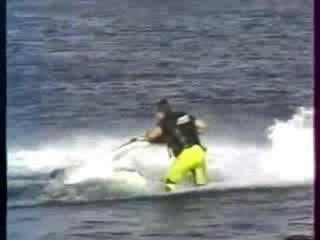} &
\includegraphics[width=0.12\textwidth]{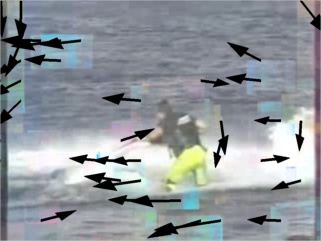} &
\includegraphics[width=0.12\textwidth]{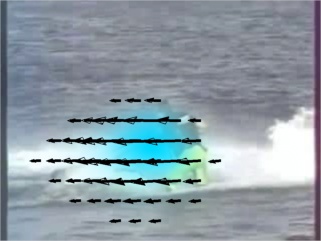} & 
\includegraphics[width=0.12\textwidth]{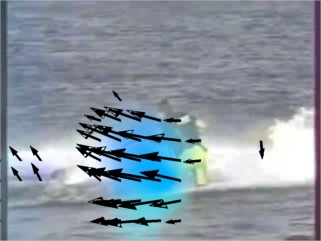} &
\includegraphics[width=0.12\textwidth]{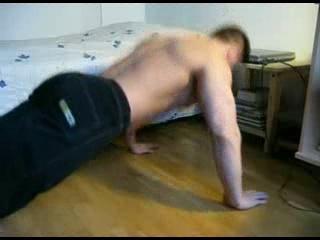} &
\includegraphics[width=0.12\textwidth]{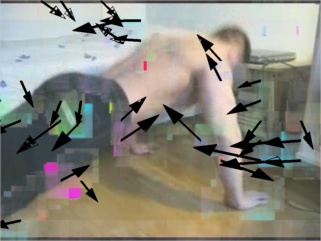} & 
\includegraphics[width=0.12\textwidth]{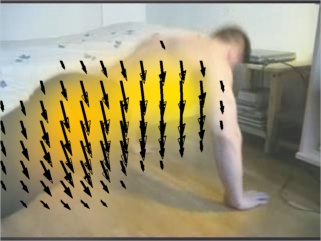} &
\includegraphics[width=0.12\textwidth]{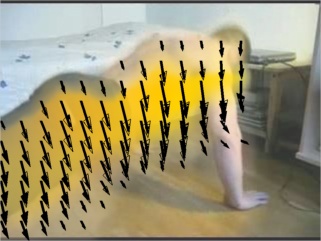} \\
\includegraphics[width=0.12\textwidth]{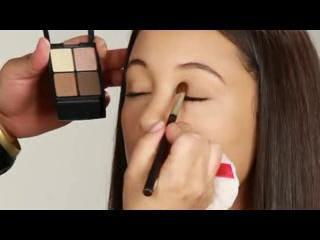} &
\includegraphics[width=0.12\textwidth]{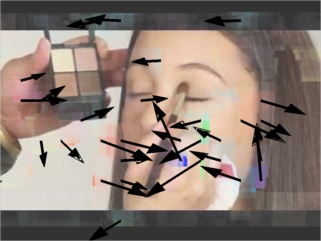} &
\includegraphics[width=0.12\textwidth]{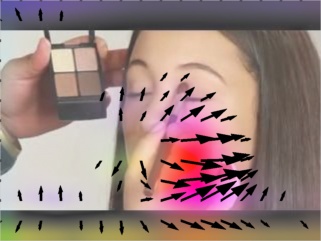} & 
\includegraphics[width=0.12\textwidth]{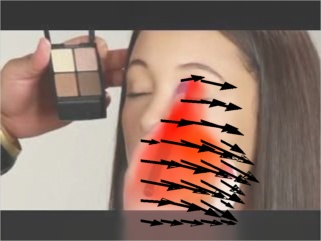} &
\includegraphics[width=0.12\textwidth]{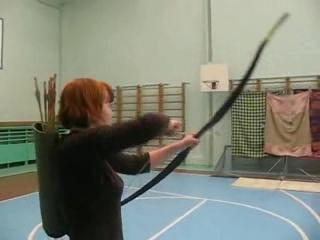} &
\includegraphics[width=0.12\textwidth]{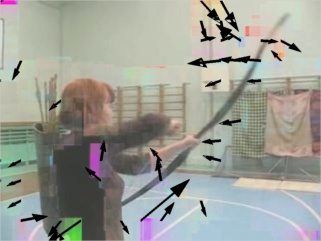} & 
\includegraphics[width=0.12\textwidth]{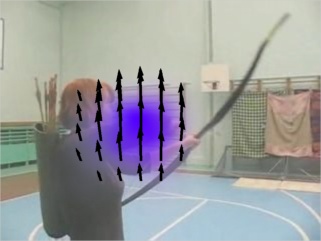} &
\includegraphics[width=0.12\textwidth]{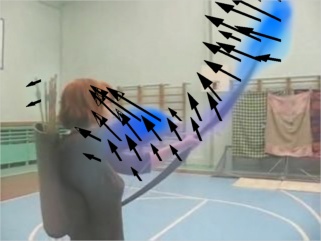} \\
\includegraphics[width=0.12\textwidth]{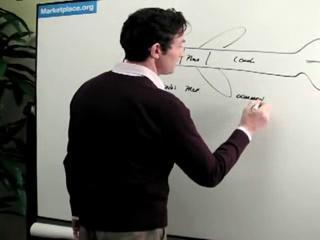} &
\includegraphics[width=0.12\textwidth]{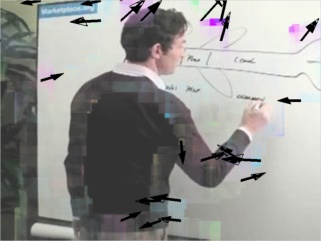} &
\includegraphics[width=0.12\textwidth]{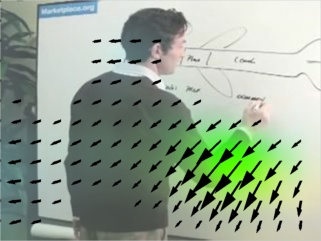} & 
\includegraphics[width=0.12\textwidth]{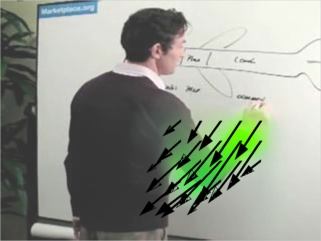} &
\includegraphics[width=0.12\textwidth]{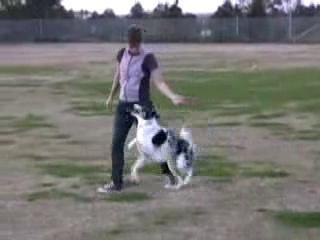} &
\includegraphics[width=0.12\textwidth]{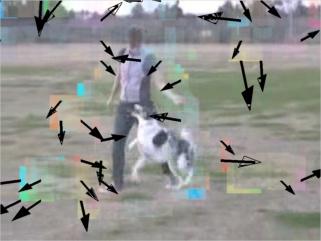} & 
\includegraphics[width=0.12\textwidth]{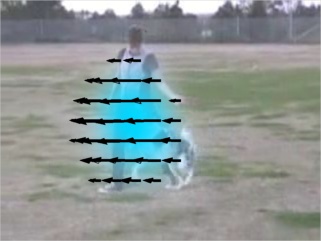} &
\includegraphics[width=0.12\textwidth]{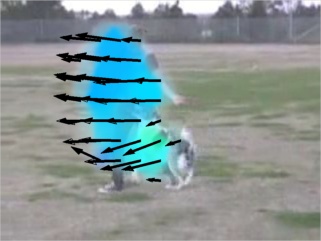} \\
\includegraphics[width=0.12\textwidth]{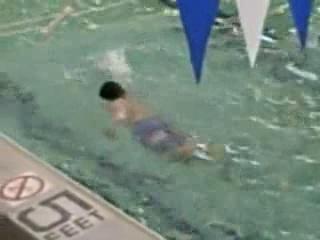} &
\includegraphics[width=0.12\textwidth]{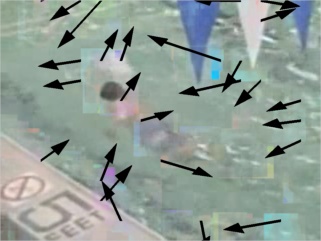} &
\includegraphics[width=0.12\textwidth]{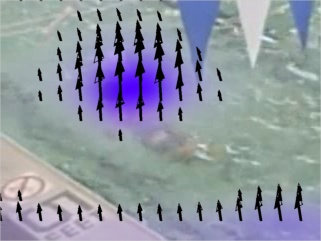} & 
\includegraphics[width=0.12\textwidth]{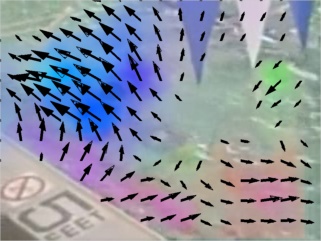} &
\includegraphics[width=0.12\textwidth]{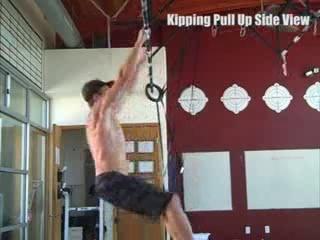} &
\includegraphics[width=0.12\textwidth]{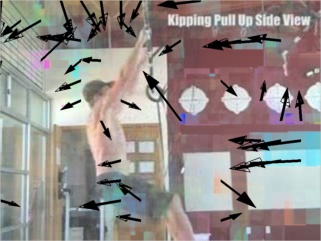} & 
\includegraphics[width=0.12\textwidth]{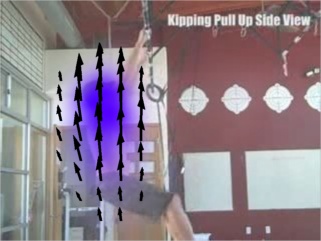} &
\includegraphics[width=0.12\textwidth]{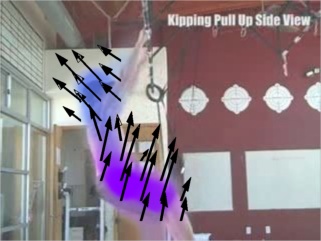} \\
{\footnotesize (a) Input Image} & {\footnotesize (b) ~\cite{Pintea14} } & {\footnotesize (c) Ours} & {\footnotesize (d) Ground Truth} &
{\footnotesize (a) Input Image} & {\footnotesize (b) ~\cite{Pintea14} }&  {\footnotesize (c) Ours} & {\footnotesize (d) Ground Truth}\\
\end{tabular}{}
\setlength{\tabcolsep}{10pt}
\begin{tabular}{ p{15cm} r }
\vspace{-0.1in}
\caption{\small Qualitative results from our method for the single frame model. While ~\cite{Pintea14} is able to predict motion in the KTH dataset (top left), we find our network strongly outperforms the baseline on more complex datasets. Our network can find the active elements in the scene and correctly predict future motion based on the context in a wide variety and scenes and actions. The color coding is on the right.}
\label{qualitative} & \raisebox{-1.0\height}{\protect\includegraphics[scale=0.4]{figures/7.jpg}}
\end{tabular}{}
\vspace{-0.5in}
\end{figure*}

For our experiments, we mostly focused on two datasets, the UCF-101 and HMDB-51, which have been popular for action recognition. For both of these datasets, we compared against baselines using 3-fold cross validation with the splits specified by the dataset organizers. We also evaluated our method on the KTH ~\cite{KTHdata} dataset using the exact same configuration in ~\cite{Pintea14} with DeepFlow. Because the KTH dataset is very small for a CNN, we finetuned our UCF-101 trained network on the training data.
For training, we subsampled frames by a factor of 5. For testing, we sampled 26,000 frames per split. For our comparison with AlexNet finetuning, we used a split which incorporated a larger portion of the training data. We will release this split publicly. 
%For our temporally deep extension, we used an additional split on each dataset which utilizes most of the data for training. We used over 60,000 frames, not overlapping in time, for the multiframe network. On the UCF-101, this meant taking out only one group per action for testing. We will release this additional split publicly. We created this split as our temporally deep model requires a large number of parameters and is more prone to overfit. 
We used three baselines for evaluation. First we used the technique of ~\cite{Pintea14}, a SRF approach to motion prediction. We took their publicly available implementation and trained a model according to their default parameters. Because of the much larger size of our datasets, we had to sample SIFT-patches less densely. We also use a Nearest-Neighbor baseline using both fc7 features from the pre-trained AlexNet network as well as pooled-5 features. Finally, we compare unsupervised training from scratch with
finetuning on the supervised AlexNet network.

\begin{table}
\centering
\small
\setlength\doublerulesep{5pt} 
\begin{tabular}{|c|c|c|c|}
\multicolumn{4}{c}{\textbf{UCF-101}}\\
\hline
Method & EPE & EPE-Canny & EPE-NZ \\ \hline
SRF ~\cite{Pintea14} & 1.30 & 1.23 & 3.24 \\ \hline
NN pooled-5 & 2.31 & 2.20 & 4.40 \\ \hline
NN fc7 & 2.24 & 2.16 & 4.27 \\ \hline
Ours-HMDB & 1.35 & 1.26 & 3.26 \\ \hline
\bfseries{Ours} & \bfseries{1.27} & \bfseries{1.17} & \bfseries{3.19} \\ \hline \hline
--- & Dir & Dir-Canny & Dir-NZ \\ \hline
SRF ~\cite{Pintea14} & .004 & .000 & -.013 \\ \hline
NN pooled-5 & -.001 & -.001  & -.067 \\ \hline
NN fc7 & -.005 & -.006 & -.060 \\ \hline
Ours-HMDB & 0.017 & 0.007 & 0.032 \\ \hline
\bfseries{Ours} & \bfseries{.045} & \bfseries{.025} & \bfseries{.092} \\ \hline \hline 
--- & Orient & Orient-Canny & Orient-NZ \\ \hline
SRF ~\cite{Pintea14} & .492 & .600  & .515 \\ \hline
NN pooled-5 & .650 & .650 & .677 \\ \hline
NN fc7 & .649 & .649 & .672 \\ \hline
Ours-HMDB & .653 & .653 & .672 \\ \hline
\bfseries{Ours} & \bfseries{.659} & \bfseries{.657} & \bfseries{.688} \\ \hline \hline
--- & Top-5 & Top-5-Canny & Top-5-NZ \\ \hline
SRF ~\cite{Pintea14} & 79.4\%  & 81.7\%  & 10.0\% \\ \hline
NN pooled-5 & 77.8\% & 79.5\% & 20.0\% \\ \hline
NN fc7 & 78.3\% & 79.9\% & 18.8\% \\ \hline
Ours-HMDB & 88.7\% & 90.0\% & 60.6\% \\ \hline
\bfseries{Ours} & \bfseries{89.7\%} & \bfseries{90.5\%} & \bfseries{65.0\%}\\ \hline \hline
--- & Top-10 & Top-10-Canny & Top-10-NZ \\ \hline
SRF ~\cite{Pintea14} &  82.2\% &  84.4\%  & 17.2\% \\ \hline
NN pooled-5 & 83.2\% & 85.3\% & 32.9\% \\ \hline
NN fc7 & 84.0\% & 85.4\% & 32.3\% \\ \hline
Ours-HMDB & 95.6\% & 95.9\% & 88.8\% \\ \hline
\bfseries{Ours} & \bfseries{96.5\%} & \bfseries{96.7\%} & \bfseries{90.9\%} \\ \hline 
\end{tabular}
\vspace{-0.1in}
\caption{\small Single-image evaluation using the 3-fold split on UCF-101. Ours-HMDB represents our network trained only on HMDB data. The Canny suffix represents pixels on the Canny edges, and the
NZ suffix represents moving pixels according to the ground-truth. NN represents a nearest-neighbor approach. Dir and Orient represent direction and orientation metrics respectively. For EPE, less is better, and for other metrics, higher is better. With the exception of Orient-NZ against
both NN features, all differences against our model are significant at the 5\% level with a paired t-test.}
\vspace{-0.2in}
\label{MetricsUCF}
\end{table}

\begin{table}
\centering
\small
\setlength\doublerulesep{5pt} 
\begin{tabular}{|c|c|c|c|}
\multicolumn{4}{c}{\textbf{HMDB-51}}\\
\hline
Method & EPE & EPE-Canny & EPE-NZ \\ \hline
SRF ~\cite{Pintea14} & 1.23 & 1.20 & 3.46 \\ \hline
NN pooled-5 & 2.51 & 2.49 & 4.89 \\ \hline
NN fc7 & 2.43 & 2.43 & 4.69 \\ \hline
Ours-UCF & 1.30 & 1.26 & 3.49 \\ \hline
\bfseries{Ours} & \bfseries{1.21} & \bfseries{1.17} & \bfseries{3.45} \\ \hline \hline
--- & Dir & Dir-Canny & Dir-NZ \\ \hline
SRF ~\cite{Pintea14} & .000 & .000 & -.010 \\ \hline
NN pooled-5 & -.008 & -.007  & -.061 \\ \hline
NN fc7 & -.007 & -.005 & -.061 \\ \hline
Ours-UCF & .016 & .011 & .003 \\ \hline
\bfseries{Ours} & \bfseries{.016} & \bfseries{.012} & \bfseries{.030} \\ \hline \hline 
--- & Orient & Orient-Canny & Orient-NZ \\ \hline
SRF ~\cite{Pintea14} & .461 & .557 & .495  \\ \hline
NN pooled-5 & .631 & .631 & .644 \\ \hline
NN fc7 & .630 & .631 & .655 \\ \hline
Ours-UCF & .634 & .634 & .664 \\ \hline
\bfseries{Ours} & \bfseries{.636} & \bfseries{.636} & \bfseries{.667} \\ \hline \hline
--- & Top-5 & Top-5-Canny & Top-5-NZ \\ \hline
SRF ~\cite{Pintea14} & 81.9\%   & 83.6\%  & 13.5\%  \\ \hline
NN pooled-5 & 76.3\% & 77.8\% & 14.0\% \\ \hline
NN fc7 & 77.3\% & 78.7\% & 13.5\% \\ \hline
Ours-UCF & 89.4\% & 89.9\% & 60.8\% \\ \hline
\bfseries{Ours} & \bfseries{90.2\%} & \bfseries{90.5\%} & \bfseries{61.0\%}\\ \hline \hline
--- & Top-10 & Top-10-Canny & Top-10-NZ \\ \hline
SRF ~\cite{Pintea14} &  84.4\%  &  86.1\%  & 22.1\%  \\ \hline
NN pooled-5 & 82.9\% & 84.0\% & 23.9\% \\ \hline
NN fc7 & 83.6\% & 84.4\% & 23.2\% \\ \hline
Ours-UCF & 95.8\% & 95.9\% & 87.6\% \\ \hline
\bfseries{Ours} & \bfseries{95.9\%} & \bfseries{95.9\%} & \bfseries{87.5\%} \\ \hline 
\end{tabular}
\vspace{-0.1in}
\caption{\small Single-image evaluation using the 3-fold split on HMDB-51. Ours-UCF represents our network trained only on UCF data. The Canny suffix represents pixels on the Canny edges, and the NZ suffix represents moving pixels according to the ground-truth. NN represents a nearest-neighbor approach. Dir and Orient represent direction and orientation metrics respectively. For EPE, less is better, and for other metrics, higher is better. With the exception of EPE-NZ against
SRF, all differences against our model are significant at the 5\% level with a paired t-test.}
\vspace{-0.1in}
\label{MetricsHMDB}
\end{table}

\begin{table}
\centering
\small
\setlength\doublerulesep{5pt} 
\begin{tabular}{ |c|c|c|c| }
\multicolumn{4}{c}{\textbf{KTH (DeepFlow)}}\\
\hline
Method & EPE & EPE-Canny & EPE-NZ\\ \hline
~\cite{Pintea14} & 0.21 & 0.19 & 1.72 \\ \hline
\bfseries{Ours} & \bfseries{0.19} & \bfseries{0.18} & \bfseries{1.17} \\ \hline \hline
--- & Orient & Orient-Canny & Orient-NZ\\ \hline
~\cite{Pintea14} & .30 & .32 & .75 \\ \hline
\bfseries{Ours} & \bfseries{.67} & \bfseries{.67} & \bfseries{.90} \\ \hline \hline
--- & Top-5 & Top-5-Canny & Top-5-NZ\\ \hline
~\cite{Pintea14} & 93.9\% & 94.4\% & 2.3\% \\ \hline
\bfseries{Ours} & \bfseries{99.0}\% & \bfseries{99.0}\% & \bfseries{98.0}\%\\ \hline 
\end{tabular}
\vspace{-0.1in}
\caption{\small We compare our network fine-tuned on the KTH dataset to ~\cite{Pintea14}. Orient represents orientation metric. NZ and Canny are non-zero and Canny pixels. All differences are significant at the 5\% level with a paired t-test.}
\vspace{-0.1in}
\label{KTH}
\end{table}

\begin{table}
\centering
\small
\setlength\doublerulesep{5pt} 
\begin{tabular}{ |c|c|c|c| }
\multicolumn{4}{c}{\textbf{Pretrained vs. From Scratch}}\\
\hline
Method & EPE & EPE-Canny & EPE-NZ\\ \hline
Pretrained & 1.19 & 1.12 & 3.12 \\ \hline
From Scratch & 1.28 & 1.21 & 3.21 \\ \hline \hline
--- & Orient & Orient-Canny & Orient-NZ\\ \hline
Pretrained & .661 & .659 & .692 \\ \hline
From Scratch & .659 & .658 & .691 \\ \hline \hline
--- & Top-5 & Top-5-Canny & Top-5-NZ\\ \hline
Pretrained & 91.0\% & 91.1\% & 65.8\% \\ \hline
From Scratch & 89.9\% & 90.3\% & 65.1\%\\ \hline 
\end{tabular}
\vspace{-0.1in}
\caption{\small We compare finetuning from ImageNet features to a randomly initialized network on UCF-101. Orient represents orientation metric. NZ and Canny are non-zero and Canny pixels.}
\vspace{-0.1in}
\label{AlexNet}
\end{table}

\begin{table}
\centering
\small
\setlength\doublerulesep{5pt} 
\begin{tabular}{ |c|c|c|c| }
\multicolumn{4}{c}{\textbf{Stabilization}}\\
\hline
Method & EPE & EPE-Canny & EPE-NZ\\ \hline
Unstabilized & 1.35 & 1.28 & 3.60 \\ \hline
Stabilized & 1.49 & 1.42 & 3.61 \\ \hline \hline
--- & Orient & Orient-Canny & Orient-NZ\\ \hline
Unstabilized & .641 & .641 & 0.664 \\ \hline
Stabilized & .652 & .652 & 0.698 \\ \hline \hline
--- & Top-5 & Top-5-Canny & Top-5-NZ\\ \hline
Unstabilized & 88.9\% & 89.2\% & 63.4\% \\ \hline
Stabilized & 88.3\% & 88.8\% & 59.7\%\\ \hline 
\end{tabular}
\vspace{-0.1in}
\caption{ \small We also compare our network trained with and without camera stabilization on a split of HMDB. Orient represents orientation metric. NZ and Canny are non-zero and Canny pixels. }
\vspace{-0.2in}
\label{Stabilization}
\end{table}

\vspace{-0.05in}
\subsection{Evaluation Metrics}
\vspace{-0.1in}
Because of the complexity and sometimes high level of label ambiguity in motion prediction, we use a variety of metrics to evaluate our method and baselines. Following from ~\cite{Pintea14}, we use traditional End-Point-Error, measuring the Euclidean distance of the estimated optical flow vector from the ground truth vector. In addition, given vectors $\mathbf{x}_{1}$ and  $\mathbf{x}_{2}$, we also measure direction similarity using the cosine similarity distance: $\frac{\mathbf{x}_{1}^{T}\mathbf{x}_{2}}{\|\mathbf{x}_{1}\|\|\mathbf{x}_{2}\|}$ and orientation similarity (angle taken on half-circle): $\frac{|\mathbf{x}_{1}^{T}\mathbf{x}_{2}|}{\|\mathbf{x}_{1}\|\|\mathbf{x}_{2}\|}$. The orientation similarity measures how  parallel is predicted optical flow vector with respect to given ground truth optical flow vector. Some motions may be strictly left-right or up-down, but the exact direction may be ambiguous. This measure accounts for this situation.

We choose these metrics established by earlier work. However, we also add some additional metrics to account for the level of ambiguity in many of the test images. As ~\cite{Pintea14} notes, EPE is a poor metric in the case where motion is small and may reasonably proceed in more than one possible direction. We thus additionally look  at the underlying distribution of the predicted classes to understand how well the algorithm accounts for this ambiguity. For instance, if we are shown an image as in Figure ~\ref{labelvariability}, it is unknown if the man will move up or down. It is certainly the case, however, that he will not move right or left.  Given the probability distribution over the quantized flow clusters, we check to see if the ground truth is within the top probable clusters. For the implementation of ~\cite{Pintea14}, we create an estimated probability distribution by quantizing the regression output from all the trees and then, for each pixel, we bin count the clusters over the trees. For Nearest-Neighbor we take the top-N matched frames and use the matched clusters in each pixel as our top-N ranking. We evaluate over the mean rank of all pixels in the image. Following ~\cite{Pintea14}, we also evaluate over the Canny edges. Because of the simplicity of the datasets in ~\cite{Pintea14}, Canny edges were a good approximation for measuring the error of pixels of moving objects in the scene. However, our data includes highly cluttered scenes that incorporate multiple non-moving objects. In addition, we find that our network is very effective at identifying moving vs non-moving elements in the scene. We find that the difference between overall pixel mean and Canny edges is very small across all metrics and baselines. Thus, we also evaluate over the moving pixels according to the ground-truth. Moving pixels in this case includes all clusters in our codebook except for the vector of smallest magnitude. While unfortunately this metric depends on the choice of codebook, we find that the greatest variation in performance and ambiguity lies in predicting the direction and magnitude of the active elements in the scene. 

\vspace{-0.05in}
\subsection{Qualitative Results}
\vspace{-0.1in}
Figure ~\ref{qualitative} shows some of our qualitative results. For single frame prediction, our network is able to predict motion in many different contexts. We find that while ~\cite{Pintea14} is able to make reasonable predictions on the KTH, qualitative performance collapses once the complexity and size of the dataset increases. Although most of our datasets consist of human actions, our model can generalize beyond simply detecting general motion on humans. Our method is able to successfully predict the falling of the ocean wave in the second row, and it predicts the motion of the entire horse in the first row. Furthermore, our network can specify motion depending on the action being performed. For the man playing guitar and the man writing on the wall, the arm is the most salient part to be moved. For the man walking the dog and the man doing a pushup, the entire body will move according to the action. 

\vspace{-0.05in}
\subsection{Quantitative Results}
\vspace{-0.05in}
\noindent {\bf UCF101 and HMDB:}
We show in tables ~\ref{MetricsUCF} and ~\ref{MetricsHMDB} that our method strongly outperforms both the Nearest-Neighbor and SRF-based baselines by a large margin on most metrics. This holds true for both datasets. Interestingly, the SRF-based approach seems to come close to ours based on End-Point-Error on all datasets, but is heavily outperformed on all other metrics. This is largely a product of the End-Point-Error metric, as we find that the SRF tends to output the mean (optical flow with very small magnitude).  This is consistent with the results found in ~\cite{Pintea14}, where actions with low, bidirectional motion can result in higher EPE than predicting no motion at all. When we account for this ambiguity in motion in the top-N metric, however, the difference in performance is large. The most dramatic differences appear over the non-zero pixels. This is due to the fact that most pixels in the image are not going to move, and an algorithm that outputs motion that is small or zero over the entire image will appear to perform artificially well without taking the moving objects into account. 

\noindent {\bf KTH:}
For KTH in table ~\ref{KTH}, ~\cite{Pintea14} is close to our method in EPE and Orientation, but Top-N suffers greatly because it often output vectors of correct direction but incorrect magnitude. On absolute levels our method seems to perform well on this simple dataset, with the network predicting the correct cluster will over $98\%$ of the time. 

\noindent {\bf Cross Dataset:} 
As both the UCF101 and HMDB dataset are curated by humans, it is important to determine how well our method is able to generalize beyond the structure of a particular dataset. In table ~\ref{MetricsUCF} we show that training on HMDB (Ours-HMDB) and testing on UCF101 leads only to a small drop in performance. Likewise, training on UCF101 (Ours-UCF101) and testing on HMDB in table ~\ref{MetricsHMDB} shows little performance loss. 

\noindent {\bf Pretraining:}
We train our representation in a self-supervised manner, using no semantic information. However, do human labels help? We compared finetuning from supervised, pretrained features trained on ImageNet to a randomly initialized network trained only on self-supervised data. The pretrained net has been exposed to far more diverse data, and the network has been trained on explicity semantic information. However, we find in table ~\ref{AlexNet} that the pretrained network yields only a very small improvement in performance. 

\noindent {\bf Stabilization:}
How robust is the network to camera motion? We explicitly stabilized the camera in our training data in order for the network to focus on moving objects and not camera motion itself. In table ~\ref{Stabilization} we compare a network trained on data with and without stabilization. We test on stablilized data, and we find even without camera stabilization that the difference in performance is small.

\vspace{-0.1in}
\section{Multi-Frame Prediction}
\vspace{-0.1in}
\begin{figure*}
\centering
\includegraphics[width=6.5in, height=2.25in]{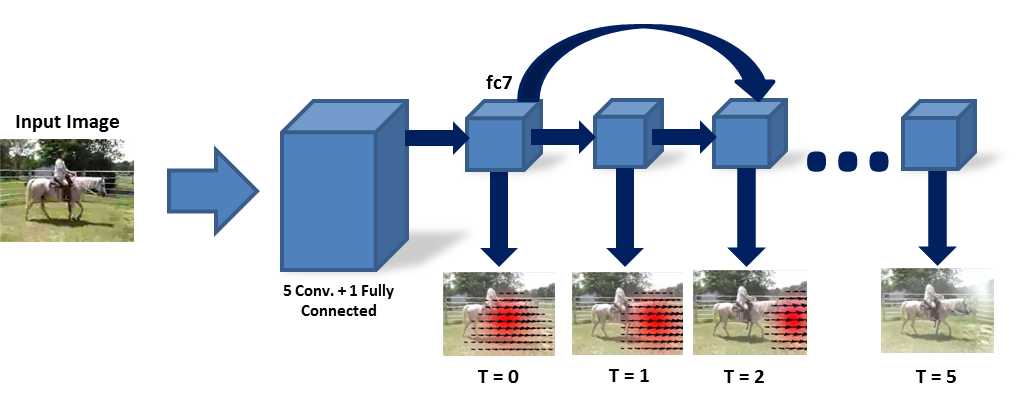} 
\vspace{-0.1in}
\caption{\small \textbf{Overview}. For our multiframe prediction, we predict entire clustered frames of optical flow as a sequence of frames. We take the learned features for our single frame model as our input, and we input them to a series of six fully connected layers, with each layer having access to the states of the past layers.}
\vspace{-0.1in}
\label{multiframe}
\end{figure*}

\begin{figure*}
\centering
\begin{tabular}{ cccccc }
\includegraphics[width=0.16\textwidth]{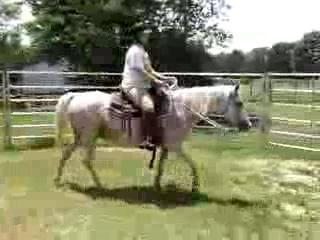} & 
\includegraphics[width=0.16\textwidth]{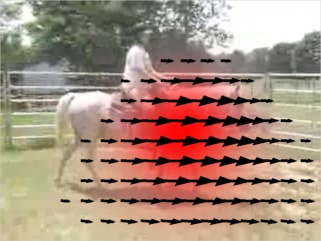} &
\includegraphics[width=0.16\textwidth]{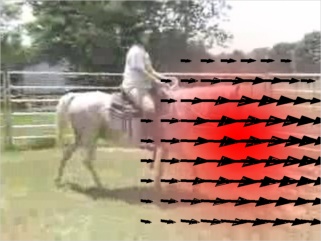} &
\includegraphics[width=0.16\textwidth]{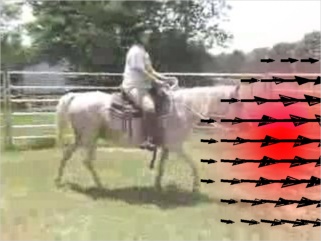} & 
\includegraphics[width=0.16\textwidth]{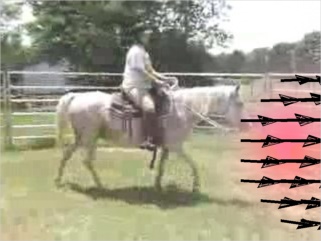} &
\includegraphics[width=0.16\textwidth]{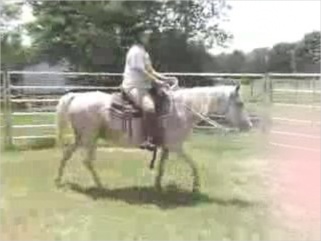} \\
\includegraphics[width=0.16\textwidth]{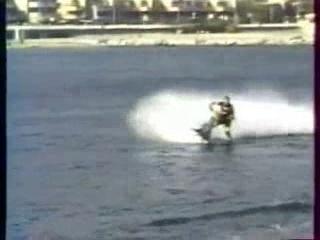} & 
\includegraphics[width=0.16\textwidth]{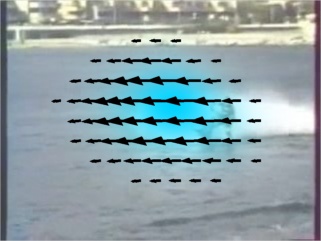} &
\includegraphics[width=0.16\textwidth]{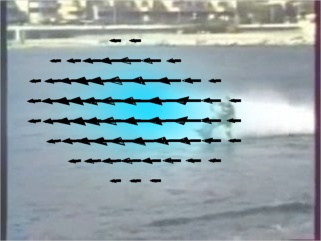} &
\includegraphics[width=0.16\textwidth]{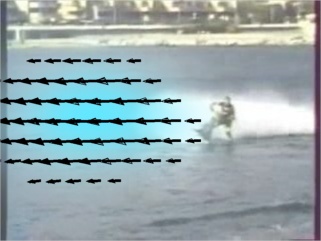} & 
\includegraphics[width=0.16\textwidth]{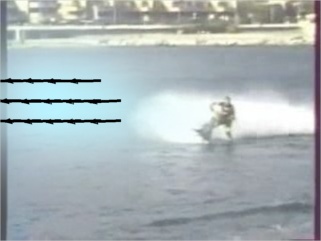} &
\includegraphics[width=0.16\textwidth]{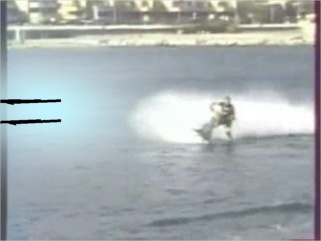} \\
\includegraphics[width=0.16\textwidth]{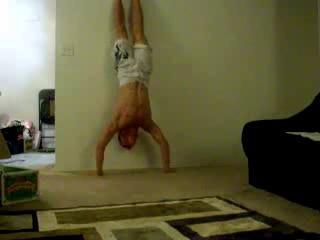} & 
\includegraphics[width=0.16\textwidth]{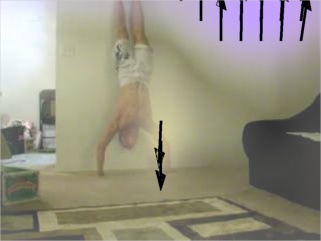} &
\includegraphics[width=0.16\textwidth]{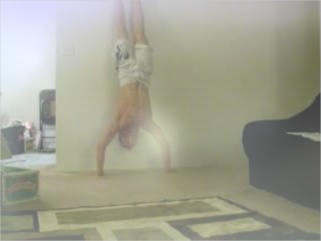} &
\includegraphics[width=0.16\textwidth]{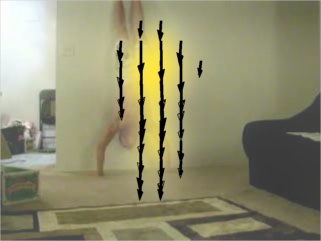} & 
\includegraphics[width=0.16\textwidth]{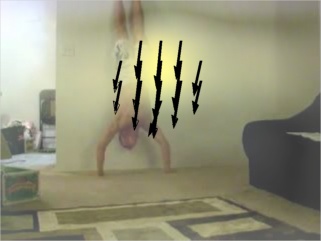} &
\includegraphics[width=0.16\textwidth]{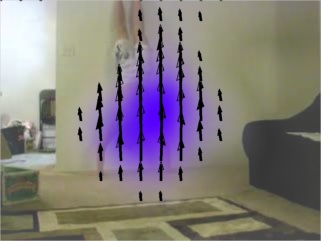} \\
\includegraphics[width=0.16\textwidth]{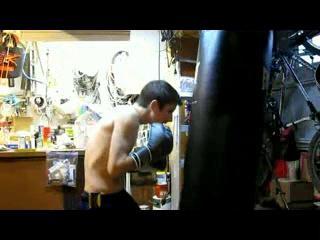} & 
\includegraphics[width=0.16\textwidth]{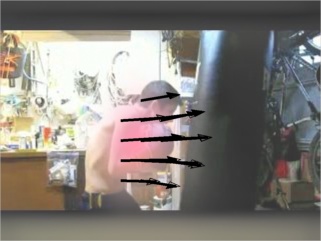} &
\includegraphics[width=0.16\textwidth]{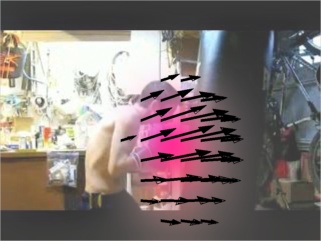} &
\includegraphics[width=0.16\textwidth]{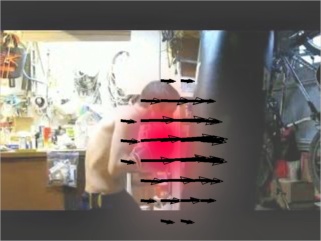} & 
\includegraphics[width=0.16\textwidth]{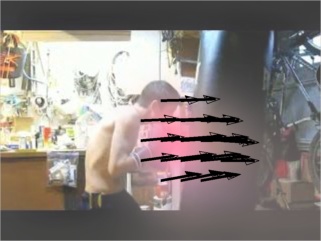} &
\includegraphics[width=0.16\textwidth]{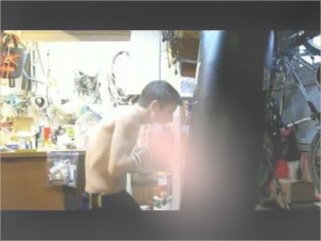} \\
{\footnotesize (a) Input Image} & {\footnotesize (b) Frame 1} & {\footnotesize (c) Frame 2} &
{\footnotesize (c) Frame 3} & {\footnotesize (d) Frame 4} & {\footnotesize (e) Frame 5}
\end{tabular}
\setlength{\tabcolsep}{10pt}
\begin{tabular}{ p{15cm} r }
\vspace{-0.1in}
\caption{\small Qualitative results for multi-frame prediction. The five rows represent predictions from our multi-frame model for future frames. Our extension can predict optical flow over multiple frames.}
\label{qualitativemulti} & \raisebox{-0.8\height}{\protect\includegraphics[scale=0.4]{figures/7.jpg}}
\end{tabular}{}
\vspace{-0.5in}
\end{figure*}

Until now we have described an architecture for predicting optical flow given a static image as input. However, it would be interesting to predict not just the next frame but a few seconds into future. How should we design such a network?

We present a proof-of-concept network to predict 6 future frames. In order to predict multiple frames into the future, we take our pre-trained single frame network and output the seventh feature layer into a "temporally deep" network, using the implementation of ~\cite{Donahue15}. This network architecture is the same as an unrolled recurrent neural network with some important differences. On a high level, our network is similar to the unfactored architecture in ~\cite{Donahue15}, with each sequence having access to the image features and the previous hidden state in order to predict the next state. We replace the LSTM module with a fully connected layer as in a RNN. However, we also do not use a true recurrent network. The weights for each sequence layer are not shared, and each sequence has access to all the past hidden states. We used 2000 hidden states in our network, but we predict at most six future sequences. We attempted to use recurrent architectures with the publicly available LSTM implementation from ~\cite{Donahue15}. However, in our experiments they always regressed to a mean trajectory across the data. Our fully connected network has much higher number of parameters than a RNN and therefore highlights the inherent difficulty of this task. Due to the much larger size of the state space, we do not predict optical flow for each and every pixel. Instead, we use kmeans to created a codebook of 1000 possible optical flow frames, and we predict one of 1000 class as output as each time step. This can be thought of as analogous to a sequential prediction problem similar to caption generation. Instead of a sequence of words, our "words" are clusters of optical flow frames, and our "sentence" is an entire trajectory. We used a set number of sequences, six, in our experiments with each frame representing the average optical flow of one-sixth of a second.

\vspace{-0.1in}
\section{Conclusion}
\vspace{-0.1in}
In this paper we have presented an approach to generalized 
prediction in static scenes. By using an optical
flow algorithm to label the data, we can train this model on a large
number of unlabeled videos. Furthermore, our framework utilizes
the success of deep networks to outperform contemporary approaches
to motion prediction. We find that our network successfully predicts motion
based on the context of the scene and the stage of the action taking place. 
Possible work includes incorporating this motion model to predict semantic action
labels in images and video. Another possible direction is to utilize the predicted optical flow to predict in raw pixel space, synthesizing a video from a single image.  

\noindent {\bf Acknowledgements:} \small We thank Xiaolong Wang for many helpful discussions.  We thank the NVIDIA Corporation for the donation of Tesla K40 GPUs for this research.  In addition, this work was supported by NSF grant IIS1227495.
 %In addition, we show we can extend our prediction further into
% the future with temporal models. 

\vspace{-0.1in}

{\small
\bibliographystyle{ieee}
\bibliography{egbib}
}

\end{document}